\documentclass[lettersize,journal]{IEEEtran}
\usepackage[numbers,sort&compress]{natbib}
\usepackage{amsmath,amsfonts}
\usepackage{algorithmic}
\usepackage{algorithm}
\usepackage{booktabs}
\usepackage{colortbl}
\usepackage{listings}
\usepackage{multirow}
\usepackage{array}
\usepackage[caption=false,font=normalsize,labelfont=sf,textfont=sf]{subfig}
\usepackage{textcomp}
\usepackage{stfloats}
\usepackage{url}
\usepackage{verbatim}
\usepackage{graphicx}
\definecolor{mygray}{rgb}{0.92,0.92,0.92}

\begin{document}

\title{EViT: An Eagle Vision Transformer with Bi-Fovea Self-Attention}

\author{Yulong Shi,~Mingwei Sun,~Yongshuai Wang,~Jiahao Ma,~Zengqiang Chen
\thanks{Manuscript received 30 April 2024; {this} work was supported in part by the National Natural Science Foundation of China under {grants} 62073177, 62473209, 61973175 and 62003351. \textit{(Corresponding author: Mingwei Sun.)}}

\thanks{Yulong Shi, Mingwei Sun, Yongshuai Wang and Jiahao Ma are with the College of Artificial Intelligence, Nankai University, Tianjin 300350, China (e-mail: ylshi@mail.nankai.edu.cn, smw\_sunmingwei@163.com, wangys@nankai.edu.cn, jhma@mail.nankai.edu.cn).}

\thanks{Zengqiang Chen is with the College of Artificial Intelligence, Nankai University, Tianjin 300350, China, and with the Key Laboratory of Intelligent Robotics of Tianjin, Tianjin 300350, China (e-mail: nkugnw@gmail.com).}}

\markboth{ }%
{}

\IEEEpubid{ }

\maketitle

\begin{abstract}

Owing to {advancements in} deep learning technology, {Vision Transformers (ViTs)} have demonstrated {impressive} performance in various computer vision tasks. {Nonetheless}, {ViTs} still face some challenges{,} such as high computational complexity and {the} absence of desirable inductive {biases}. To alleviate these issues, {the potential advantages of combining eagle vision with ViTs are explored. We summarize a Bi-Fovea Visual Interaction (BFVI) structure inspired by the unique physiological and visual characteristics of eagle eyes. A novel Bi-Fovea Self-Attention (BFSA) mechanism and Bi-Fovea Feedforward Network (BFFN) are proposed based on this structural design approach, which can be used to mimic the hierarchical and parallel information processing scheme of the biological visual cortex, enabling networks to learn feature representations of targets in a coarse-to-fine manner. Furthermore, a Bionic Eagle Vision (BEV) block is designed as the basic building unit based on the BFSA mechanism and BFFN. By stacking BEV blocks, a unified and efficient family of pyramid backbone networks called Eagle Vision Transformers (EViTs) is developed.} Experimental results show that EViTs exhibit highly competitive performance in various computer vision tasks{,} such as image classification, object detection and semantic segmentation. {Compared with other approaches,} EViTs {have} significant advantages{, especially in terms of performance and computational efficiency}. {The developed code} is available at \url{https://github.com/nkusyl/EViT}

\end{abstract}

\begin{IEEEkeywords}
Bi-Fovea Self-Attention, {Bi-Fovea Feedforward Network}, Eagle Vision Transformer.
\end{IEEEkeywords}

\section{Introduction}

\IEEEPARstart{O}{ver} {the past decade, Convolutional Neural Networks (CNNs) have made extraordinary contributions to the field of computer vision in accordance with the development of deep learning technologies and hardware computing units~\cite{liu2022convnet, yao2023dual, liu2024vmamba}. This success can be attributed to the pyramidal structural design of CNNs and their inherent inductive biases, such as translation invariance and local sensitivity.} However, it is {challenging} for CNNs to {explicitly model the importance of global contextual information because of} the limited receptive fields {possessed by} convolutional kernels. This issue restricts the further development and applications of CNNs~\cite{wang2023crossformer++, papa2024survey}. {Moreover}, {inspired by the success of Transformers~\cite{choi2024efficient, chen2024survey} in the Natural Language Processing (NLP) field, a question has been raised: What happens when transformers are applied in the field of computer vision? The original Vision Transformer (ViT)~\cite{dosovitskiy2020image} is a significant milestone that first incorporated a Transformer into vision tasks, which demonstrates superior performance in computer vision tasks, {and have attracted} widespread attention worldwide~\cite{han2022survey, liu2023survey}.}

\begin{figure}[t]
	\centering
	\begin{tabular}{c}
		\hspace{-3mm}\includegraphics[width=0.5\textwidth]{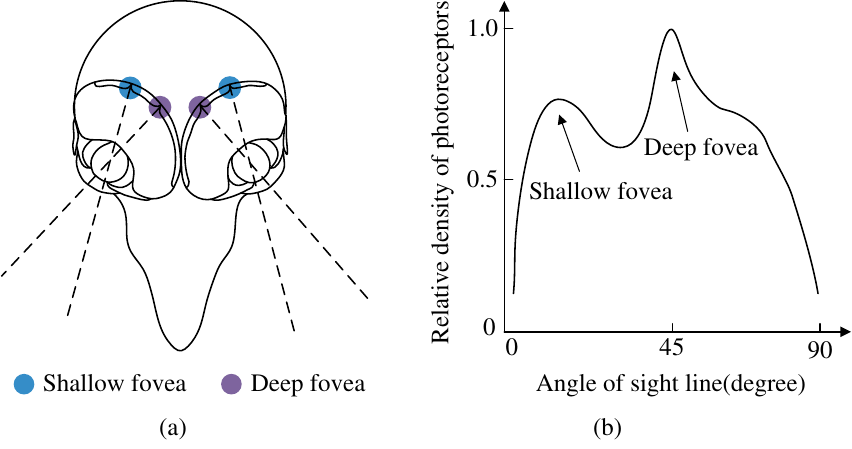}
	\end{tabular}
	\caption{(a) The physiological structure of the bi-fovea in eagle eye; (b) The density distribution of {the} photoreceptor cells {contained} in {the} bi-fovea.}
	\label{fig:eagle_eyes}
\end{figure} 

{Following the release of the ViT~\cite{dosovitskiy2020image}}, various vision transformer variants~\cite{wang2024repvit,zhang2024vision, xiao2024ctnet} have been proposed, offering new paradigms and solutions for {completing} computer vision tasks~\cite{rao2023tgfuse, tang2022matr, ma2023visualizing}. Nonetheless, {ViTs} also face several challenges, including {the following.} {(1) The process of computing self-attention matrices in ViTs has proven to be their major efficiency bottleneck. This is because calculating the importance of each token in an input sequence to the other tokens is necessary, which leads to quadratic computational complexity with respect to the length of the sequence~\cite{wang2022pvt, liu2023survey, li2023bvit}. Consequently, the maximum input sequence length is limited by the amount of available memory. This issue is especially notable when addressing high-resolution images and videos.} (2) {ViTs} tend to focus on the overall information {of objects when processing target features, thus lacking sensitivity to local features and details. This limitation impairs their performance in dense prediction tasks. (3) Due to the absence of appropriate inductive biases, ViTs require more data for training. In scenarios with limited available data, this increases the risk of overfitting.}

To alleviate these issues, we draw inspiration from {biological} vision and {develop} a unified pyramid backbone network. Figure~\ref{fig:eagle_eyes} {illustrates} the physiological structure and photoreceptor cell density distribution {of} the bi-fovea of eagle eyes. Although eagle vision and vision transformer {originate from biological science and computer science, respectively}, {we identify three analogous properties between them.} (1) Attention {mechanisms}: Eagle vision is renowned for its fast {focusing ability, which enables eagles to efficiently capture prey} in complex environments. Similarly, {the self-attention mechanism} in vision transformers allows the network to dynamically allocate attention scores to different regions, {enabling them to encode the critical} feature representations of targets. (2) {Multilevel} feature extraction {capabilities}: Eagles process visual information at multiple levels, {beginning with} photoreceptor cells and ultimately reaching {the} cerebral cortex. {In the same way, vision transformers encode} the target features {in a} layer-by-layer {manner} by stacking the Multi-Head Self-Attention (MHSA) and a Multi-Layer Perceptron (MLP). (3) Global information awareness: Eagle vision {involves} a wide field of view {for perceiving prey and predators at high altitudes and from long distances.} Furthermore, vision transformers can {encode the target features observed across the spatial domain of the entire input image.} This global perception ability allows vision transformers to {model and integrate the} contextual information {among different locations within an} image, which helps {them obtain a} more accurate understanding of the image and {process it better.}

\begin{figure*}[t]
	\centering         
	\includegraphics[width=\textwidth]{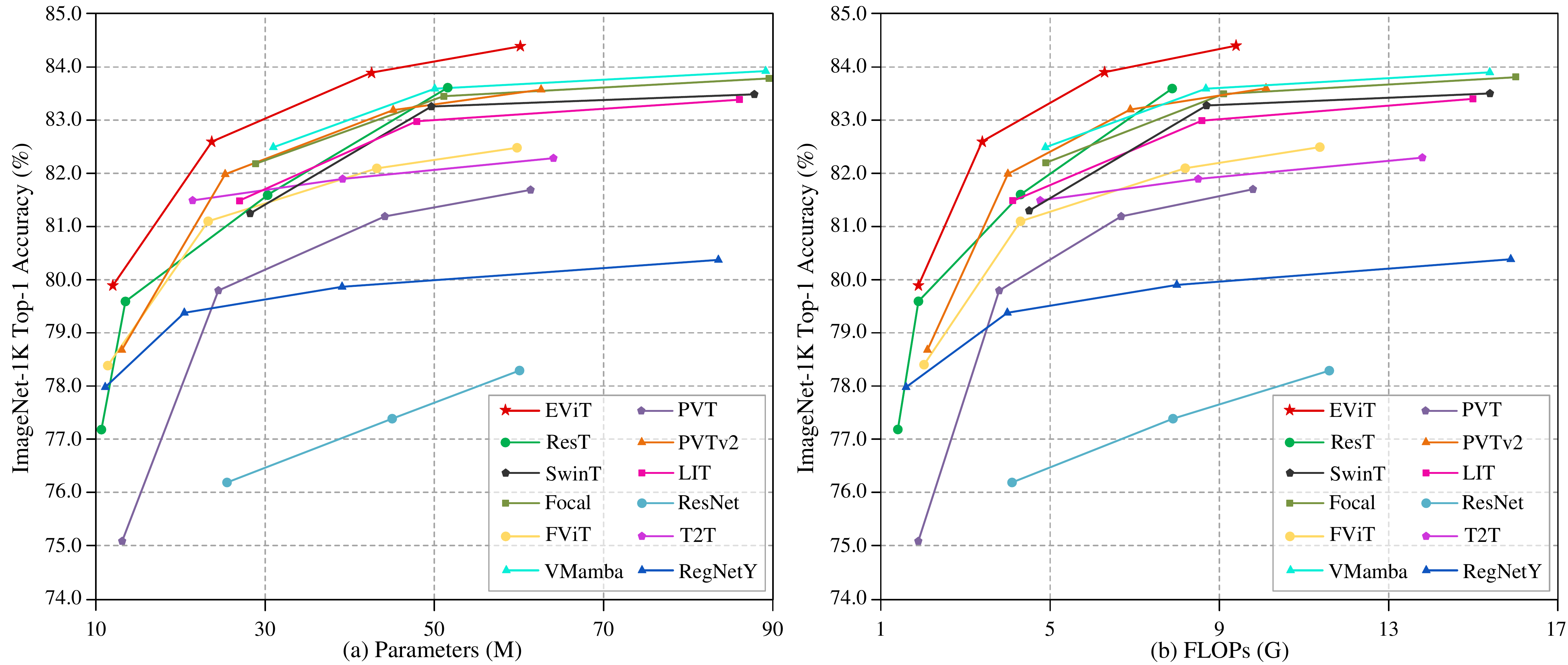}
	\caption{Comparison {among the Top-1 accuracies achieved by the} EViTs and other baselines on the ImageNet-1K dataset. {Compared with their counterparts,} the EViTs achieve better {trade-offs among the number of parameters}, {the} computational complexity and performance.}
	\label{fig:flops_pamer}
\end{figure*} 

According to the above discussions, we revisit the potential advantages of combining eagle vision with vision transformer. {A Bi-Fovea Visual Interaction (BFVI) structure is derived from the unique physiological and visual characteristics of eagle eyes, which can integrate the benefits of both cascade and parallel structures, including hierarchical organization and parallel information processing. Thereafter, a novel Bi-Fovea Self-Attention (BFSA) mechanism and a Bi-Fovea Feedforward Network (BFFN) are proposed based on this design approach, which mimics the hierarchical and parallel information processing scheme of the biological visual cortex. As improved variants of self-attention and feedforward networks{, respectively, the} BFSA and BFFN can be used to help networks extract features in a coarse-to-fine manner, resulting in high computational efficiency and scalability. Furthermore, a Bionic Eagle Vision (BEV) block is designed as the basic building unit based on BFSA and BFFN. Following the mainstream hierarchical design concept~\cite{wang2022pvt, yao2023dual, zhu2023biformer}, a general family of pyramid vision backbone networks called Eagle Vision Transformers (EViTs) is developed by stacking BEV blocks. These} EViTs {include} four variants{, namely,} EViT-Tiny, EViT-Small, EViT-Base and EViT-Large, enhancing {their} applicability in various computer vision tasks. Figure~\ref{fig:flops_pamer} {presents} the {results of a} performance comparison {between the} EViTs {and} other convolutional neural networks and vision transformers on the ImageNet~\cite{russakovsky2015imagenet} dataset. To our knowledge, this is the first work to {integrate} eagle vision with vision transformers on large-scale datasets such as ImageNet~\cite{russakovsky2015imagenet}, and {it} is also the first study to propose a general {family of} vision backbone networks {on the basis of} eagle vision.

The main contributions {of this study} are as follows.

\begin{itemize}
	\setlength\itemsep{0em}
	\item Benefiting from biological vision, {a structural design approach called Bi-Fovea Vision Interaction (BFVI) is derived from the unique physiological and visual characteristics of eagle eyes, which integrates the advantages of both cascade and parallel connections, featuring hierarchical organization and parallel information processing.}
	
	\item {Based on the BFVI design approach, a novel Bi-Fovea Self-Attention (BFSA) and a Bi-Fovea Feedforward Network (BFFN) are proposed to construct a Bionic Eagle Vision (BEV) block. These components are used to simulate the shallow and deep fovea of eagle vision, enabling the network to learn the feature representations of the target from coarse-to-fine.}
	
	\item Following the hierarchical design concept, {a general and efficient family of pyramid backbone networks called EViTs is developed. Experimental results demonstrate that the EViTs perform excellently across various vision tasks and exhibit significant competitive advantages in terms of computational efficiency and scalability.}
\end{itemize}

The remainder of this paper is structured as follows. Section~2 summarizes the related work {on} biological eagle vision and vision transformer, respectively. Section~3 describes the process of {designing} EViTs. Section~4 {presents} the experimental results {produced by the} EViTs {in} various vision tasks. Section~5 {presents} the conclusion.

\section{Related Work}

\subsection{Biological Eagle Vision}

Eagles possess {an} excellent natural visual system, {allowing them to observe their surroundings with remarkable sensitivity, which is attributed to their unique bi-fovea visual structure.} {The physiological structure and photoreceptor cell density distribution of eagle eyes {are illustrated} in Figure~\ref{fig:eagle_eyes}. The} deep fovea is located at the centre of the retina and has a high photoreceptor cells density. This is crucial for enhancing the visual resolution of eagle eyes, {allowing them to recognize and capture prey at long distances~\cite{gonzalez2017visual}.} The shallow fovea is located in the peripheral area of the retina, {where the photoreceptor cell density is lower,} but it provides a wider field of view~\cite{wu2024biological, li2022biological}. {This structure not only enhances the efficiency of eagles in terms of processing visual information, but also ensures their survival in complex environments.}

\begin{figure}[bp]
	\centering
	\begin{tabular}{c}
		\hspace{-3mm}\includegraphics[width=0.5\textwidth]{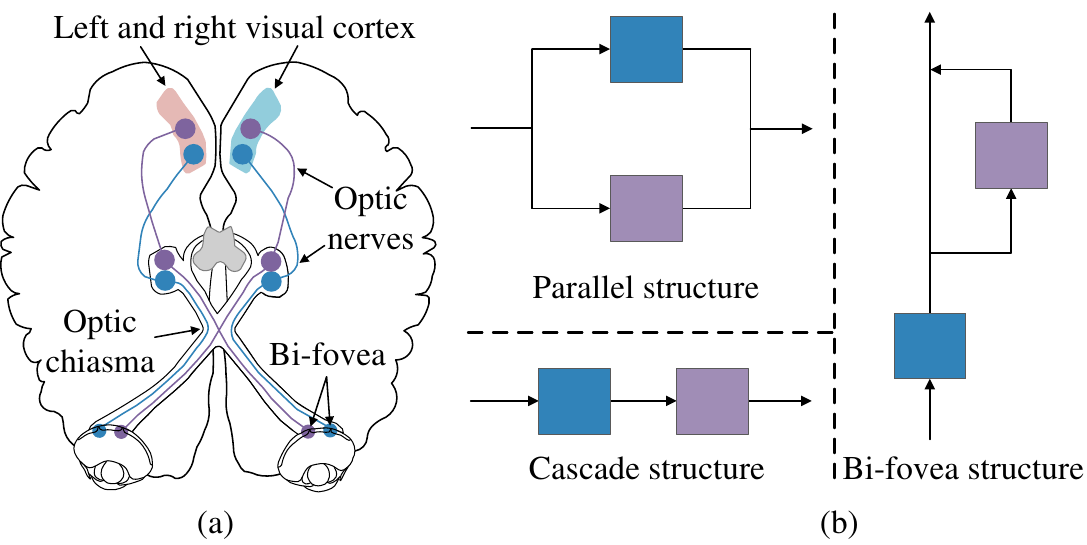}
	\end{tabular}
	\caption{(a) The visual neural pathway of the eagle eyes; (b) The Bi-Fovea Visual Interaction (BFVI) structure.}
	\label{fig:nearal_path}
\end{figure}

{Figure~\ref{fig:nearal_path}(a) illustrates the visual neural pathway of an eagle. As mentioned above, eagles process visual information at multiple levels, beginning with photoreceptor cells and ultimately reaching the cerebral cortex. Although one eagle eye cannot simultaneously use both the deep and shallow fovea for imaging, two eyes can collaborate to alternate between them~\cite{bringmann2019structure, guzman2021deepfoveanet}.} For example, when an eagle is looking forward, the deep fovea of one eye can be used for fine target recognition, {whereas} the shallow fovea of the other eye can be used to perceive the surrounding {environment}. {Inspired by the unique physiological and visual characteristics of eagle eyes, a Bi-Fovea Vision Interaction (BFVI) structure is summarized in this work. The BFVI structure, which is illustrated in Figure~\ref{fig:nearal_path}(b), integrates the benefits of cascaded and parallel structures, including hierarchical organization and parallel information processing.}

\subsection{Transformers for Vision}

ViT~\cite{dosovitskiy2020image} is a pioneering work that first introduces transformers into vision tasks. {The core idea is to split images into non-overlapping patches and then feed them as sequences into self-attention mechanisms and multilayer perceptrons for feature encoding.} Following {the} ViT, a series of improvement methods {have} been proposed~\cite{liu2021swin, touvron2021training, fan2024kmt}. {The ViT is extended in various directions, including self-attention mechanisms~\cite{fan2024lightweight, yao2023dual}, positional encoding~\cite{shaw2018self, wang2022convolutional}, and computational efficiency~\cite{zhang2021rest, liu2023efficientvit}. Compared with CNNs, vision transformers are excellent at modelling long-range dependencies and capturing global contextual information~\cite{gu2022multi, papa2024survey}. These {strengths enable vision transformers to exhibit potential surpassing that of CNNs in vision tasks such as} image classification~\cite{xue2022deep, bhojanapalli2021understanding, zhang2022evolutionary}, object detection~\cite{zhao2024ms, pu2024rank, xie2023vit}, and semantic segmentation~\cite{hsu2023video, shen2024mtc, zheng2023knowledge, gao2021novel}.} {The success of vision transformers can be attributed primarily to their innovative architectural designs, which include self-attention mechanisms, multi-layer perceptrons, and skip connections. These components allow networks to learn the interactions between the spatial locations of features in raw data.} 
	
{The CF-ViT~\cite{chen2023cf} is a coarse-to-fine visual neural network that achieves satisfactory performance. The difference between this network and our method is that we do not employ the two-stage approach for network inference. Instead, we draw inspiration from eagle vision to enhance the ability of our model to focus on image details and the global context. This is beneficial for the expansion and application of EViTs across various visual tasks. Moreover, vision mamba models~\cite{pei2024efficientvmamba, shi2024multi, liu2024vmamba}, which are state-space models with linear computational complexity, have also demonstrated tremendous potential in representation learning tasks, gaining significant attention because of their efficiency and ability to model long-term dependencies in sequences. MLLA~\cite{han2024demystify} explores the relationship between state-space models and linear attention~\cite{katharopoulos2020transformers}, reinterpreting vision mamba as a variant of the linear attention transformer. Representative vision mamba models include EfficientVMamba~\cite{pei2024efficientvmamba}, MSVMamba~\cite{shi2024multi}, and VMamba~\cite{liu2024vmamba}. In the experiments, these models are compared with EViTs in detail.} Our work demonstrates the potential of combining eagle vision with vision transformers, showing that EViTs can {lead to} more performance breakthroughs in vision tasks.

\begin{figure*}[t]
	\centering         
	\includegraphics[width=\textwidth]{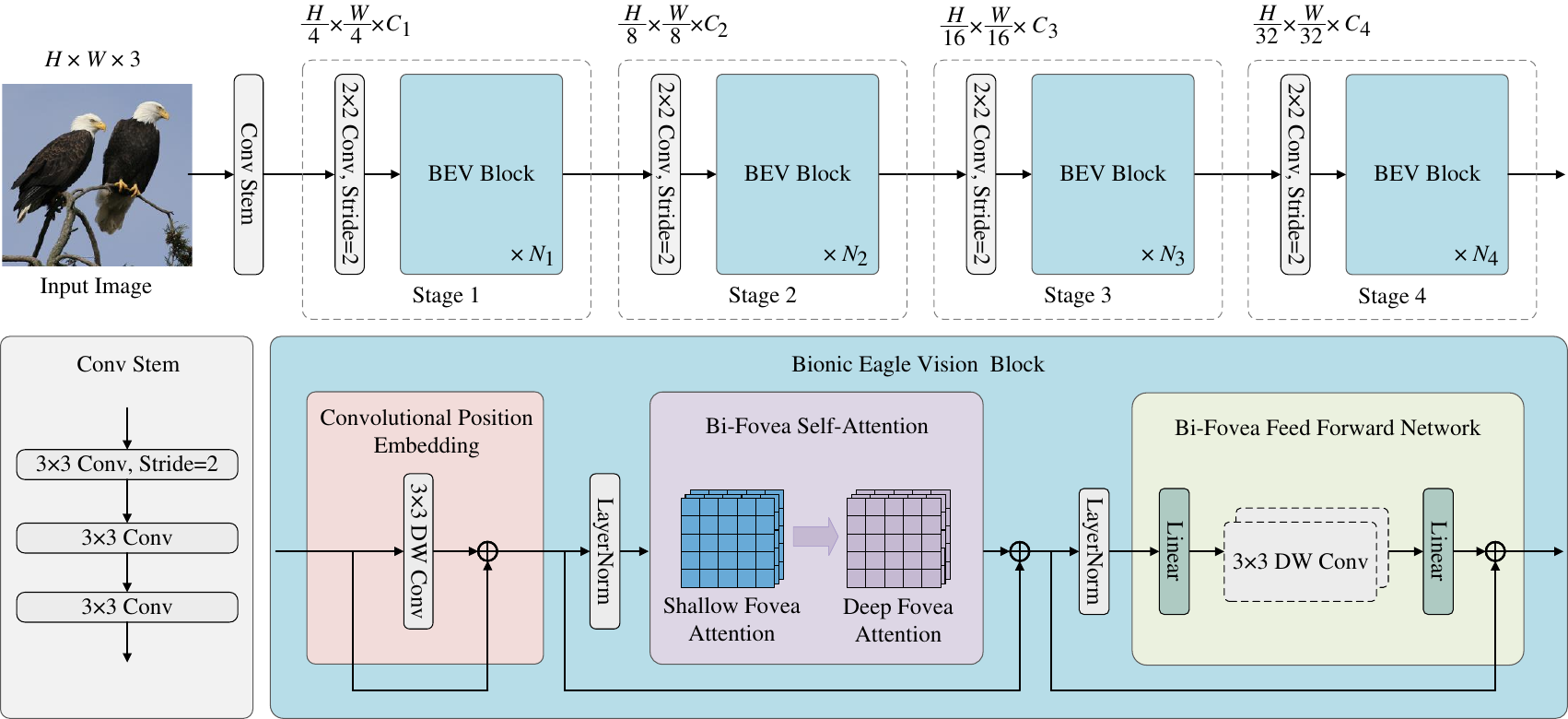}
	\caption{Illustration of the EViT. {The} EViT is composed of a convolutional stem and a pyramid structure with four stages. Each stage includes a 2 × 2 convolution with {a stride of} 2 and multiple Bionic Eagle Vision (BEV) blocks. {Each} BEV block consists of three key components: a Convolutional Positional Embedding (CPE), a Bi-Fovea Self-Attention (BFSA) and a Bi-Fovea Feedforward Network (BFFN).}
	\label{fig:EViT}
\end{figure*} 

\section{Approach}

\subsection{Overall Architecture}

{Inspired by} eagle vision, {a novel family of pyramid backbone networks called Eagle Vision Transformers (EViTs) is proposed.} The overall pipeline of {an} EViT{, which} is illustrated in Figure~\ref{fig:EViT}{, consists of a convolutional stem, several $2 \times 2$ convolutional layers and Bionic Eagle Vision (BEV) blocks. Here, the stride of these $2 \times 2$ convolutional layers is 2, and the layers are used for patch embedding.} As general backbone {networks for various vision} tasks, EViTs follow the mainstream hierarchical design concept~\cite{wang2021pyramid, yao2023dual, zhu2023biformer}{, which comprises four stages, each of which possesses a similar architecture.} The difference is that the {resolutions} of the features {output} from stage 1 to stage 4 {are} divided by factors of 4, 8, 16 and 32, respectively, {while} the corresponding channel dimensions are increased to $C_1$, $C_{2}$, $C_{3}$ and $C_{4}$. Given an input image of size $H \times W \times 3$, it is first fed into the convolutional stem to obtain the low-level feature representations. {The convolutional stem consists of three successive $3 \times 3$ convolutional layers,} where the first {convolutional} layer {has a stride of} 2{, which} is used to stabilize the training process in the early stages of the network. {These} low-level representations are {then} processed through a series of $2 \times 2$ {convolutional} layers and BEV blocks to generate hierarchical representations of {the} targets. Finally, in image classification tasks, a normalization layer, an average pooling layer, and a fully connected layer are used as classifiers to output the predictions.

\subsection{Bionic Eagle Vision Block}

As the basic building {units} of EViTs, BEV blocks {integrates the advantages} of convolutions and vision transformers. A BEV block consists of three key components: a Convolutional Position Embedding (CPE), a Bi-Fovea Self-Attention (BFSA) and a Bi-Fovea Feedforward Network (BFFN). The complete mathematical definition of {a} BEV block is shown as

\begin{equation}
	\begin{aligned}\!\!\!\!\!\!\!\!\!\!\!\!\!
		{\bf{X}} = {\rm{CPE}}({{\bf{X}}_{in}}) + {{\bf{X}}_{in}}
	\end{aligned}
	\label{eq:(1)}
\end{equation}

\begin{equation}
	\begin{aligned}\!\!\!\!\!\!
		{\bf{Y}} = {\rm{BFSA}}({\rm{LN(}}{\bf{X}}{\rm{)}}) + {\bf{X}}
	\end{aligned}
	\label{eq:(2)}
\end{equation}

\begin{equation}\!\!\!\!\!\!
	\begin{aligned}
		{\bf{Z}} = {\rm{BFFN}}({\rm{LN}}({\bf{Y}})) + {\bf{Y}}
	\end{aligned}
	\label{eq:(3)}
\end{equation}

\noindent{where}, LN represents the {layer normalization} function, which is used to normalize the feature tensors. {Stage} 1 {is taken} as an example. Given an input tensor ${{\bf{X}}_{in}} \in {\mathbb{R}^{\frac{H}{4} \times \frac{W}{4} \times {C_1}}}$, it is first processed by the CPE, which is used to introduce position information into all tokens. {The structure of the CPE is shown in Figure 4, and it is defined as}

\begin{equation}\!\!\!\!\!\!
	\begin{aligned}
		{{\rm{CPE}}({{\rm\bf{X}}_{in}}) = {\rm{DWConv}}({{\rm\bf{X}}_{in}})}
	\end{aligned}
	\label{eq:(21)}
\end{equation}

{The position information plays a crucial role in describing visual representations. In previous works, most vision transformer models encode position information through Absolute Position Embedding (APE)~\cite{liu2021swin} and Relative Position Embedding (RPE)~\cite{shaw2018self}. These embeddings are typically defined by a series of sinusoidal functions with varying frequencies or learnable parameters. The difference is that APE is designed for specific input sizes and is not robust to changes in the resolution of feature tokens, requiring adjustment to accommodate these variations. On the other hand, the distances among the feature tokens in an input sequence are considered by RPE, which exhibits translational invariance. However, additional computational costs are introduced when calculating the relative distances among feature tokens. Importantly, visual tasks still require absolute position information, and RPE cannot provide this information.} 
	
{For the above reasons, Convolutional Position Embedding (CPE) is introduced in EViTs to encode the position information of feature tokens, which is composed of a depth-wise convolution. Compared with APE and RPE, there are two advantages of CPE as follows. (1) CPE can flexibly learn the position information of features at arbitrary resolutions through zero padding with a depth-wise convolution, providing it with plug-and-play properties. (2) CPE further integrates the necessary inductive bias into EViTs through a depth-wise convolution, which helps improve their performance ceiling. Additionally}, this BEV block employs BFSA to simulate the shallow fovea and deep fovea of eagle vision for {modelling} the global feature dependencies and local fine-grained feature representations {of} images. Finally, {the BFFN is used to} complement the local information, and to improve the information interaction and local feature extraction {capabilities of the} BEV blocks.

\begin{figure}[t]
	\centering\small\resizebox{1.0\linewidth}{!}{
		\begin{tabular}{c}
			\hspace{-3mm}\includegraphics[width=0.515\textwidth]{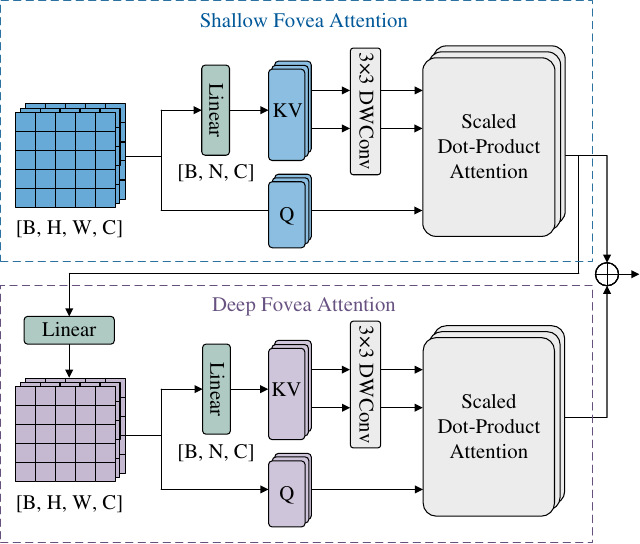}
	\end{tabular}}
	\caption{Illustration of the BFSA. The BFSA consists of a Shallow Fovea Attention (SFA) and a Deep Fovea Attention (DFA)}
	\label{fig:BFSA}
\end{figure}

\subsection{Bi-Fovea Self-Attention}

{Eagle eyes possess unique bi-fovea physiological and visual characteristics.} The shallow fovea of eagle vision is {utilized} for coarse-grained environmental perception, and the deep fovea is used for fine-grained prey recognition. {Inspired by} this fact, a Bi-Fovea Self-Attention (BFSA) is proposed. {An} illustration of this BFSA is shown in Figure~\ref{fig:BFSA}. The BFSA consists of a Shallow Fovea Attention (SFA) and a Deep Fovea Attention (DFA). In terms of {the} structural design {of BFSA}, {the SFA and DFA are not simply connected in parallel or in cascade. Instead, the Bi-Fovea Vision Interaction (BFVI) structure derived from eagle vision is adopted.} This BFVI {structural} combines the advantages of both parallel and {cascaded} connections, {allowing the SFA to model the global feature dependencies of images while enabling the DFA to capture the fine-grained feature representations of the targets.} 

\noindent\textbf{Shallow Fovea Attention.} Given an input tensor ${{\bf{X}}} \in {\mathbb{R}^{{H} \times {W} \times {C}}}$, it is first projected {to a} Query ${\bf{Q}} \in {\mathbb{R}^{N \times C}}$, {a} Key ${\bf{K}} \in {\mathbb{R}^{N \times C}}$ and {a} Value ${\bf{V}} \in {\mathbb{R}^{N \times C}}$. {Intuitively, adopting MHSA for the feature maps to calculate the global dependencies among tokens is a redundant and time-consuming approach. To mitigate the computational complexity and memory requirements of SFA, DWConv is employed to reduce the lengths of ${\bf{K}}$ and ${\bf{V}}$. The dimensionality of ${\bf{Q}}$ remains unchanged, which directly affects the ability of self-attention to focus on input the features. Reducing the dimensionality of ${\bf{Q}}$ can degrade the ability to fully capture the information possessed by the input, thereby impacting the effectiveness of the attention mechanism.} Specifically, ${\bf{Q'}} = {\rm{Linear(}}{{\bf{X}}}{\rm{)}}$, ${\bf{K'}} = {\rm{Linear}}({\rm{DWConv}}({{\bf{X}}}))$ and ${\bf{V'}} = {\rm{Linear}}({\rm{DWConv}}({{\bf{X}}}))$. {This is different from the designs of LSRA~\cite{wang2022pvt} and SRA~\cite{wang2021pyramid}, which use average pooling and standard convolution, respectively. Compared with LSRA and SRA, our design achieves a better trade-off between reducing the computational complexity of the network and preserving the independence of spatial and channel features. Furthermore, SFA is used to encode the global feature dependencies among all tokens to produce attention scores.} The compact matrix form of the SFA is defined as

\begin{equation}
	\begin{aligned}
		{\rm{SFA}}({{\bf{X}}}) = {\rm{Concat}}({\bf{hea}}{{\bf{d}}_0},{\bf{hea}}{{\bf{d}}_1},...,{\bf{hea}}{{\bf{d}}_h}){\bf{W}}
	\end{aligned}
	\label{eq:(4)}
\end{equation}

\begin{equation}
	\begin{aligned} 
		{\bf{hea}}{{\bf{d}}_i} = {\rm{Attention(}}{{\bf{Q'}}_i}{\rm{,}}{{\bf{K'}}_i},{{\bf{V'}}_i}{\rm{)}}
	\end{aligned}
	\label{eq:(5)}
\end{equation}

\begin{equation}
	\begin{aligned}\!\!\!
		{\rm{Attention}}({\bf{Q'}},{\bf{K'}},{\bf{V'}}) = {\rm{softmax}}(\frac{{{\bf{Q'}}{{{\bf{K'}}}^\top}}}{{\sqrt D }}){\bf{V'}}
	\end{aligned}
	\label{eq:(6)}
\end{equation}

\noindent{where} ${\bf{hea}}{{\bf{d}}_i} \in {\mathbb{R}^{N \times \frac{D}{h}}}$ is the output of the $i^{th}$ attention head, and the weight matrix ${{\bf{W}}}\in {\mathbb{R}^{N \times \frac{D}{h}}}$ is used to {construct all the} heads. 

\begin{figure}[t]
	\centering\small\resizebox{1.0\linewidth}{!}{
		\begin{tabular}{c}
			\hspace{-3mm}\includegraphics[width=0.515\textwidth]{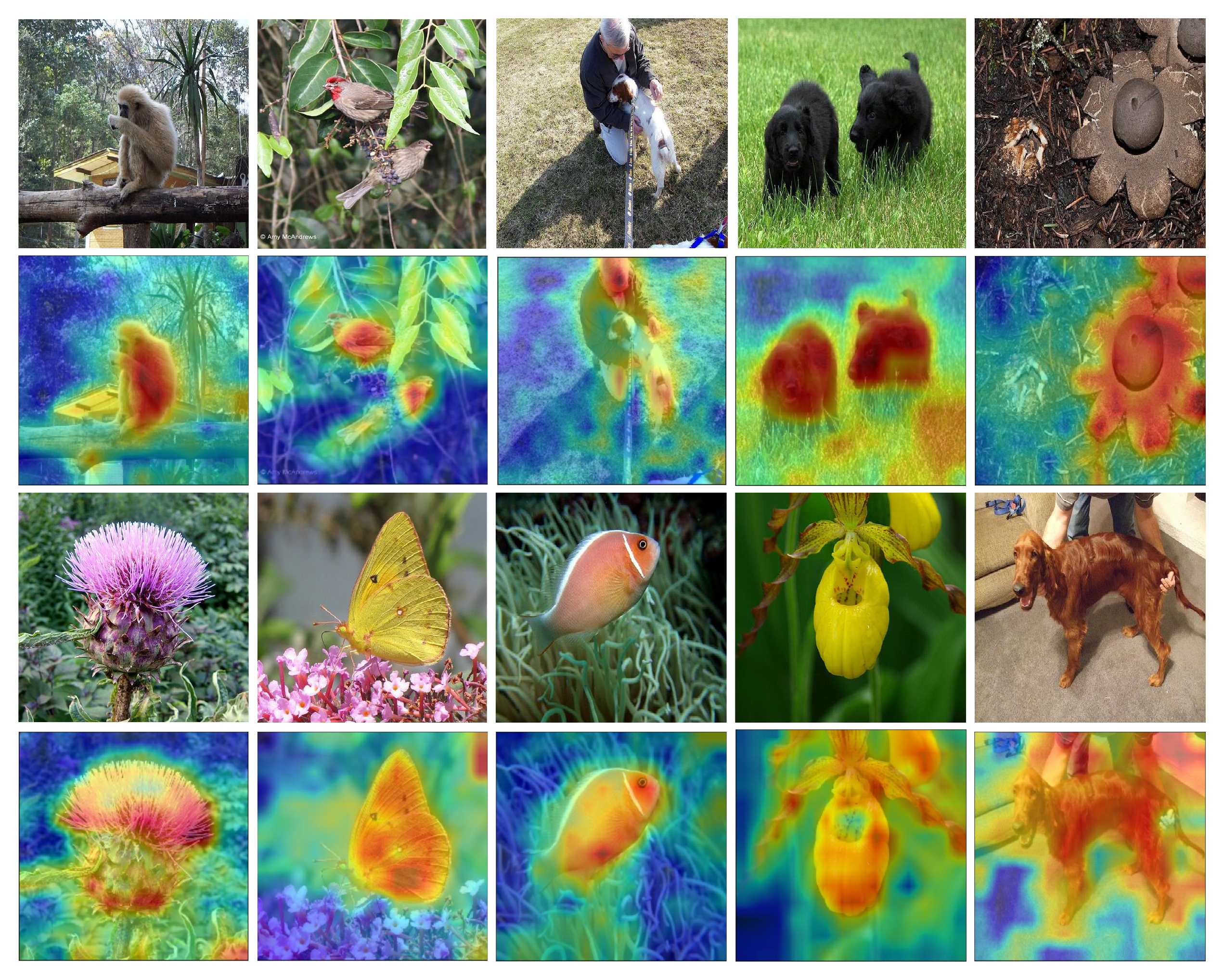}
	\end{tabular}}
	\caption{The attention map of {an} EViT. The BFSA {pays more} attention to the foreground targets of interest}
	\label{fig:visual}
\end{figure}

\begin{table*}[bp]
	\caption{Four architectural EViT {variants} for ImageNet classification. $H_i$ denotes the number of attention heads {contained in the} DFA and SFA of stage \emph{i}. $c_i$ and $f_i$ are used to control the reduced {sizes of the} feature tokens {in} stage \emph{i}. $r_i$ denotes the expansion ratio {of the} BFFN {in} stage \emph{i}.}
	\renewcommand{\arraystretch}{0.88}
	\setlength\tabcolsep{5pt}
	\centering
	\small
	\begin{tabular}{c|c|c|c|c|c}
		\toprule [0.125em]
		Output size & Layer Name & EViT-Tiny & EViT-Small & EViT-Base & EViT-Large \\ \midrule
		$112\times112$ & Conv Stem
		& $\begin{array}{c} 3\times3, 28, \text{stride}~2 \\ \left[3\times3, 28\right] \times 2 \end{array}$
		& $\begin{array}{c} 3\times3, 32, \text{stride}~2 \\ \left[3\times3, 32\right] \times 2 \end{array}$
		& $\begin{array}{c} 3\times3, 32, \text{stride}~2 \\ \left[3\times3, 32\right] \times 2 \end{array}$
		& $\begin{array}{c} 3\times3, 36, \text{stride}~2 \\ \left[3\times3, 36\right] \times 2 \end{array}$ \\ \midrule
		
		$56\times56$ & Patch Embedding & $2\times2$, $56$, stride $2$ & $2\times2$, $64$, stride $2$ & $2\times2$, $64$, stride $2$ & $2\times2$, $72$, stride $2$ \\ \midrule
		
		\begin{tabular}{c} Stage 1 \end{tabular} & \begin{tabular}{c}BEV block\end{tabular}
		& $\begin{bmatrix}\setlength{\arraycolsep}{1pt} \begin{array}{c}
				H_1$=$1, k$=$8\\ f$=$8, r_1$=$3
		\end{array} \end{bmatrix} \times 2$
		& $\begin{bmatrix}\setlength{\arraycolsep}{1pt} \begin{array}{c}
				H_1$=$1, k$=$8\\ f$=$8, r_1$=$3
		\end{array} \end{bmatrix} \times3$
		& $\begin{bmatrix}\setlength{\arraycolsep}{1pt} \begin{array}{c}
				H_1$=$2, k$=$8\\ f$=$8, r_1$=$3.5
		\end{array} \end{bmatrix} \times 4$
		& $\begin{bmatrix}\setlength{\arraycolsep}{1pt} \begin{array}{c}
				H_1$=$2, k$=$8\\ f$=$8, r_1$=$4
		\end{array} \end{bmatrix} \times 4$ \\ \midrule
		
		$28\times28$ & Patch Embedding & $2\times2$, $112$, stride $2$ & $2\times2$, $128$, stride $2$ & $2\times2$, $128$, stride $2$ & $2\times2$, $144$, stride $2$ \\ \midrule 
		\begin{tabular}{c} Stage 2 \end{tabular} & \begin{tabular}{c}BEV block \end{tabular} 
		& $\begin{bmatrix}\setlength{\arraycolsep}{1pt} \begin{array}{c}
				H_2$=$2, k$=$4\\ f$=$4, r_2$=$3
		\end{array} \end{bmatrix} \times 4$
		& $\begin{bmatrix}\setlength{\arraycolsep}{1pt} \begin{array}{c}
				H_2$=$2, k$=$4\\ f$=$4, r_2$=$3
		\end{array} \end{bmatrix} \times5$
		& $\begin{bmatrix}\setlength{\arraycolsep}{1pt} \begin{array}{c}
				H_2$=$4, k$=$4\\ f$=$4, r_2$=$3.5
		\end{array} \end{bmatrix} \times 8$
		& $\begin{bmatrix}\setlength{\arraycolsep}{1pt} \begin{array}{c}
				H_2$=$4, k$=$4\\ f$=$4, r_2$=$4
		\end{array} \end{bmatrix} \times 8$ \\ \midrule
		
		$14\times14$ & Patch Embedding & $2\times2$, $224$, stride $2$ & $2\times2$, $256$, stride $2$ & $2\times2$, $256$, stride $2$ & $2\times2$, $288$, stride $2$ \\ \midrule
		
		\begin{tabular}{c} Stage 3 \end{tabular} & \begin{tabular}{c}BEV block
		\end{tabular} 
		& $\begin{bmatrix}\setlength{\arraycolsep}{1pt} \begin{array}{c}
				H_3$=$4, k$=$2\\ f$=$2, r_3$=$3
		\end{array} \end{bmatrix} \times 8$
		& $\begin{bmatrix}\setlength{\arraycolsep}{1pt} \begin{array}{c}
				H_3$=$4, k$=$2\\ f$=$2, r_3$=$3
		\end{array} \end{bmatrix} \times15$
		& $\begin{bmatrix}\setlength{\arraycolsep}{1pt} \begin{array}{c}
				H_3$=$8, k$=$2\\ f$=$2, r_3$=$3.5
		\end{array} \end{bmatrix} \times 27$
		& $\begin{bmatrix}\setlength{\arraycolsep}{1pt} \begin{array}{c}
				H_3$=$8, k$=$2\\ f$=$2, r_3$=$4
		\end{array} \end{bmatrix} \times 30$ \\ \midrule
		
		$7\times7$ & Patch Embedding & $2\times2$, $448$, stride $2$ & $2\times2$, $512$, stride $2$ & $2\times2$, $512$, stride $2$ & $2\times2$, $576$, stride $2$ \\ \midrule
		
		\begin{tabular}{c} Stage 4 \end{tabular} & \begin{tabular}{c}BEV block
		\end{tabular} 
		& $\begin{bmatrix}\setlength{\arraycolsep}{1pt} \begin{array}{c}
				H_4$=$8, k$=$1\\ f$=$1, r_4$=$3
		\end{array} \end{bmatrix} \times 2$
		& $\begin{bmatrix}\setlength{\arraycolsep}{1pt} \begin{array}{c}
				H_4$=$8, k$=$1\\ f$=$1, r_4$=$3
		\end{array} \end{bmatrix} \times3$
		& $\begin{bmatrix}\setlength{\arraycolsep}{1pt} \begin{array}{c}
				H_4$=$16, k$=$1\\ f$=$1, r_4$=$3.5
		\end{array} \end{bmatrix} \times 4$
		& $\begin{bmatrix}\setlength{\arraycolsep}{1pt} \begin{array}{c}
				H_4$=$16, k$=$1\\ f$=$1, r_4$=$4
		\end{array} \end{bmatrix} \times 4$ \\ \midrule
		
		$1\times1$ & Projection & \multicolumn{4}{c}{$1\times1$, $1280$} \\ \midrule
		$1\times1$ & Classifier & \multicolumn{4}{c}{Fully Connected Layer, $1000$} \\ \midrule
		\multicolumn{2}{c|}{Params} & $11.8$ M & $24.9$ M & $42.8$ M & $61.9$ M \\ \midrule
		\multicolumn{2}{c|}{FLOPs} & $2.1$ G & $4.1$ G & $7.2$ G & $10.3$ G \\
		\bottomrule[0.125em]
	\end{tabular}
	
	\label{tab:arch_variants_of_EviT}
\end{table*}

\noindent\textbf{Deep Fovea Attention.} The mathematical definitions of DFA and SFA are {identical}, and the difference is {that the output of SFA serves as the input for DFA. This design approach fully utilizes the advantages of self-attention, enhancing the ability of the network to extract complex features through a layer-by-layer refinement process.} The complete mathematical definition of {the} DFA is shown as

\begin{equation}\!\!\!\!\!
	\begin{aligned}
		{\rm{DFA}}({\bf{X'}}) = {\rm{Concat}}({\bf{hea}}{{\bf{d}}'_0},{\bf{hea}}{{\bf{d}}'_1},...,{\bf{hea}}{{\bf{d}}'_h}){\bf{W'}}
	\end{aligned}
	\label{eq:(7)}
\end{equation}

\noindent{where}

\begin{equation}\!\!\!\!\!\!
	\begin{aligned}
		{\bf{X'}} = {\rm{SFA}}({{\bf{X}}})
	\end{aligned}
	\label{eq:(8)}
\end{equation}

Finally, {the outputs of the SFA and DFA are summed and fed into the next layer. This design follows the visual characteristics of the information interaction and fusion processes implemented in the bi-fovea of eagle eyes, and is used to simulate the working mechanisms of information processing in biological visual systems. The process is represented as}

\begin{equation}\!\!\!\!\!\!\!
	\begin{aligned}
		{\bf{Out}} = {\rm{SFA(}}{{\bf{X}}}{\rm{) + DFA(}}{{\bf{X'}}}{\rm{)}}
	\end{aligned}
	\label{eq:(9)}
\end{equation}

{To conduct an in-depth analysis of the BFSA, the activation mapping approach is employed to generate visual attention maps, which are shown in Figure~\ref{fig:visual}, wherein the BFSA allocates varying attention scores to different regions and targets while processing image features. Regions with higher attention values are crucial for the current task, reflecting the feature selection priorities of the network. According to the numerical distribution of these visual attention maps, the BFSA pays more attention to the foreground objects of interest, effectively concentrating on the task-relevant features while suppressing unnecessary background information. This improves the performance of the model, providing richer feature information for visual tasks.}

\subsection{Complexity Analysis of {the} BFSA}

{In this subsection, a computational complexity analysis is conducted on the standard MHSA and BFSA. For simplicity, the attention heads are excluded. Given input feature maps with dimensions of $h \times w \times d$, the computational complexity of the standard MHSA is $3hw{d^2} + 2{h^2w^2}d$, where $3hw{d^2}$ represents the computational complexity of generating ${\bf{Q}}$, ${\bf{K}}$ and ${\bf{V}}$, which involves three linear projections. $2{h^2w^2}d$ denotes the complexity of computing the attention weights, which includes matrix multiplying ${\bf{Q}}$ and ${\bf{K}^\top}$, along with a Softmax operation. The computational complexity of self-attention is quadratic with respect to the sequence length. In SFA, the computational complexity of the linear projections of ${\bf{Q}}$, ${\bf{K}}$ and ${\bf{V}}$ is $3n{d^2}$, which is consistent with that of MHSA. The difference is that SFA uses a DWConv to reduce the spatial sizes of ${\bf{K}}$ and ${\bf{V}}$. The computational complexity of this operation is $hwc$. The computational complexity of the attention matrix and the softmax operation in SFA is $2{h^2w^2}d/{f^2}$, where $f$ is the reduction factor for the spatial sizes of ${\bf{K}}$ and ${\bf{V}}$. Overall, the computational complexity of this SFA is $3hw{d^2} + hwc + 2{h^2w^2}d/{f^2}$. The computational complexity of BFSA is twice that of SFA. Since the sequence length of the feature tokens is reduced by the BFSA through depth-wise convolution, its overall computational complexity and memory cost remain low, even though self-attention calculations are performed twice. Additionally, an inductive bias is further introduced to the EViT by using DWConv to process ${\bf{K}}$ and ${\bf{V}}$. This step enhances the ability of the network to model local feature information, which helps to improve the robustness and generalizability of the network in complex scenarios.}

\subsection{Bi-Fovea Feedforward Network}

\begin{figure}[t]
	\centering\small\resizebox{1.0\linewidth}{!}{
		\begin{tabular}{c}
			\hspace{-3mm}\includegraphics[width=0.515\textwidth]{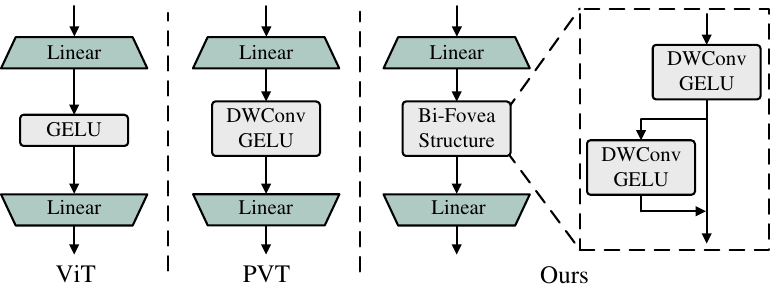}
	\end{tabular}}
	\caption{{Structural} illustration of the Bi-Fovea Feedforward Network (BFFN). {The} FFN in {the} ViT~\cite{dosovitskiy2020image} (left), {the} CFFN in PVT v2~\cite{wang2022pvt} (right), and our BFFN {are compared}.}
	\label{fig:BFFN}
\end{figure}

As essential {components} of {vision} transformers, {the feedforward networks are employed to enhance the representation of image features through nonlinear transformations.} However, {the feedforward network lacks sensitivity to local features.} The common practice is to introduce {convolutional} operations between two {fully} connected layers or use $1 \times 1$ convolutions {to replace the fully connected layers, which is considered inefficient.} To this end, we are inspired by the ability of the biological visual cortex to process information and believe that an efficient feed forward network should satisfy the imposed hierarchical structure and parallel information processing requirements. {Therefore, the design approach of the BFVI structure is followed, and a Bi-Fovea Feedforward Network (BFFN) is proposed.} The structure of the BFFN is illustrated in Figure~\ref{fig:BFFN}. As {already emphasized}, {the} BFFN {combines the characteristics of the} hierarchical structure and parallel information processing, which {helps to} increase the receptive field of each network layer and improve the multi-scale feature {representations} of networks with finer {granularity. In {our} experiments, ablation studies are conducted on the connection structure and effectiveness of the BFFN.}

\subsection{{Architectural} Variants of EViTs}

The EViTs {include} four variations: EViT-Tiny, EViT-Small, EViT-Base and EViT-Large. {These variants have different parameter scales and computational complexity. To obtain multiscale feature representations, each stage of these variants uses a 2 × 2 convolution with a stride of 2 for connection purposes. This convolution is applied for patch embedding. At the beginning of the next stage, the spatial {sizes of the} feature maps are halved, and the number of dimensions is doubled.} {Therefore, the four stages of EViTs can output feature maps with four different sizes, providing rich hierarchical feature representations of the targets. These hierarchical representations are beneficial for enhancing the performance of the EViTs in dense prediction tasks.} The configuration details of {the} EViTs are shown in Table~\ref{tab:arch_variants_of_EviT}. To facilitate comparison with other mainstream networks, the input image resolutions of EViT-Tiny, EViT-Small, EViT-Base and EViT-Large are all ${224^2}$.

\begin{table}[t]
	\caption{ImageNet-1K classification results {obtained by the} EViTs. We groups similar CNNs and vision transformers together based on {their} model parameters and classification performance.}
	\begin{center}
		\resizebox{1.0\linewidth}{!}{
			\begin{tabular}{l|c|c|c|c}
				\toprule[0.125em]
				\multirow{1}{*}{Model} & Resolution & FLOPs (G)& Params (M)& Top-1 Acc (\%)\\
				
				\midrule
				ResNet-18~\cite{he2016deep} & $224^2$ & 1.8 & 11.7 & 69.8 \\
				PVT-T~\cite{wang2021pyramid} & $224^2$ & 1.9 & 13.2 & 75.1 \\
				EfficientVMamba-T~\cite{pei2024efficientvmamba} & $224^2$ & {0.8} & {6.0} & 76.5 \\
				ResT-Lite~\cite{zhang2021rest} & $224^2$ & 1.4 & 10.5 & 77.2 \\
				RegNetY-1.6~\cite{radosavovic2020designing} & $224^2$ & 1.6 & 11.2 & 78.0 \\
				FViT-T~\cite{shi2024fvit} & $224^2$ & 2.0 & 11.7 & 78.4 \\
				ResT-Small~\cite{zhang2021rest} & $224^2$ & 1.9 & 13.6 & 79.6 \\
				PVTv2-B1~\cite{wang2022pvt} & $224^2$ & 2.1 & 13.1 & 78.7 \\
				EfficientVMamba-S~\cite{pei2024efficientvmamba} & $224^2$ & 1.3 & 11.0 & 78.7 \\
				MSVMamba-M~\cite{shi2024multi} & $224^2$ & 1.5 & 12.0 & 79.8 \\
				RepViT-M1.5~\cite{wang2024repvit} & $224^2$ & 2.3 & 14.0 & {81.2} \\
				\rowcolor{mygray} EViT-Tiny & $224^2$ & 2.1 & 11.8 & 79.9 \\
				
				\midrule
				ResNet-50~\cite{he2016deep} & $224^2$ & 4.1 & 25.6 & 76.2 \\
				RegNetY-4.0~\cite{radosavovic2020designing} & $224^2$ & 4.0 & {20.6} & 79.4 \\
				PVT-S~\cite{wang2021pyramid} & $224^2$ & 3.8 & 24.5 & 79.8 \\
				FViT-S~\cite{shi2024fvit} & $224^2$ & 4.3 & 23.4 & 81.1 \\
				Swin-T~\cite{liu2021swin} & $224^2$ & 4.5  & 28.3 & 81.3 \\
				T2T-14~\cite{yuan2021tokens} & $224^2$ & 5.2 & 21.5 & 81.5 \\
				LIT-S~\cite{pan2022less} & $224^2$ & 4.1 & 27.0 & 81.5 \\
				ResT-Base~\cite{zhang2021rest} & $224^2$ & 4.3 & 30.3 & 81.6 \\
				PVTv2-B2~\cite{wang2022pvt} & $224^2$ & 4.0 & 25.4 & 82.0 \\
				ConvNeXt-T~\cite{liu2022convnet} & $224^2$ & 4.5 & 28.0 & 82.1 \\
				Focal-T~\cite{yang2021focal} & $224^2$ & 4.9 & 29.1 & 82.2 \\
				RepViT-M2.3~\cite{wang2024repvit} & $224^2$ & 4.5 & 22.9 & 82.5 \\
				VMamba-T~\cite{liu2024vmamba} & $224^2$ & 4.9 & 30.0 & {82.6} \\
				\rowcolor{mygray} EViT-Small & $224^2$ & {4.1} & 24.9 & {82.6}\\
				
				\midrule
				ResNet-101~\cite{he2016deep} & $224^2$ & 7.9 & 44.7 & 77.4 \\
				RegNetY-8.0~\cite{radosavovic2020designing} & $224^2$ & 8.0 & 39.2 & 79.9 \\
				PVT-M~\cite{wang2021pyramid} & $224^2$ & 6.7 & 44.2 & 81.2 \\
				EfficientVMamba-B~\cite{pei2024efficientvmamba} & $224^2$ & 4.0 & {33.0} & 81.8 \\
				T2T-19~\cite{yuan2021tokens} & $224^2$ & 9.8 & 39.2 & 81.9 \\
				FViT-B~\cite{shi2024fvit} & $224^2$ & 8.2 & 43.3 & 82.1 \\
				MSVMamba-M~\cite{shi2024multi} & $224^2$ & {4.6} & {33.0} &  82.8 \\
				LIT-M~\cite{pan2022less} & $224^2$ & 8.6 & 48.0 & 83.0 \\
				Swin-S~\cite{liu2021swin} & $224^2$ & 8.7 & 49.6 & 83.0 \\
				ConvNeXt-S~\cite{liu2022convnet} & $224^2$ & 8.7 & 50.0 & 83.1 \\
				PVTv2-B3~\cite{wang2022pvt} & $224^2$ & 6.9 & 45.2 & 83.2 \\
				Focal-S~\cite{yang2021focal} & $224^2$ & 9.1 & 51.1 & 83.5 \\
				ResT-Large~\cite{zhang2021rest} & $224^2$ & 7.9 & 51.6 & 83.6 \\
				VMamba-S~\cite{liu2024vmamba} & $224^2$ & 8.7 & 50.0 & 83.6 \\
				\rowcolor{mygray} EViT-Base & $224^2$ & 7.2 & 42.8 & {83.9} \\
				
				\midrule
				ResNet-152~\cite{he2016deep} & $224^2$ & 11.6 & 60.2 & 78.3 \\
				RegNetY-16~\cite{radosavovic2020designing} & $224^2$ & 15.9 & 83.6 & 80.4 \\
				PVT-L~\cite{wang2021pyramid} & $224^2$ & 9.8 & 61.4 & 81.7 \\
				T2T-24~\cite{yuan2021tokens} & $224^2$ & 15.0 & 64.1 & 82.2 \\
				FViT-L~\cite{shi2024fvit} & $224^2$ & 11.4 & 59.8 & 82.5 \\
				LIT-B~\cite{pan2022less} & $224^2$ & 15.0 & 86.0 & 83.4 \\
				Swin-B~\cite{liu2021swin} & $224^2$ & 15.4 & 87.8 & 83.5 \\
				PVTv2-B4~\cite{wang2022pvt} & $224^2$ & 10.1 & 62.6 & 83.6 \\
				Focal-B~\cite{yang2021focal} & $224^2$ & 16.0 & 89.8 & 83.8 \\
				ConvNeXt-B~\cite{liu2022convnet} & $224^2$ & 15.4 & 89.0 & 83.8 \\
				VMamba-B~\cite{liu2024vmamba} & $224^2$ & 15.4 & 89.0 & 83.9 \\
				\rowcolor{mygray} EViT-Large & $224^2$ & {10.3} & {61.9} & {84.4}\\
				\rowcolor{mygray} EViT-Large & $256^2$ & {13.7} & {61.9} & {84.9}\\
				\bottomrule[0.125em]
		\end{tabular}}
	\end{center}
	\label{tab:class_results}
\end{table}

\section{Experiments}

In this section, experiments are conducted with {the} EViTs on a series of mainstream computer vision tasks, including ImageNet-1K~\cite{russakovsky2015imagenet} classification ({Section} 4.1), COCO 2017~\cite{lin2014microsoft} object detection and instance segmentation ({Section} 4.2), ADE20K~\cite{zhou2017scene} semantic segmentation  ({Section} 4.3), and other transfer learning tasks ({Section} 4.4). Specifically, the EViTs are trained first from scratch on {the} ImageNet-1K dataset to implement image classification and obtain {their pretraining} parameters. {The pretraining} parameters are {subsequently} fine-tuned on object detection, semantic segmentation and other vision tasks through transfer learning, which is {done} to validate the generalization performance of {the} EViTs.  {Additionally, {a} robustness evaluation and {an} ablation study are conducted {on} EViTs in Sections 4.5 and 4.6{, respectively,} to demonstrate the effectiveness of {the} BFSA and BFFN.}

\subsection{Image Classification on ImageNet-1k}

\noindent\textbf{Settings.} In this section, the EViTs are first evaluated on the ImageNet-1K~\cite{russakovsky2015imagenet} dataset, which {includes} 1000 classes and {approximately} 1.33M images in total. Among these, the training dataset contains {approximately} 1.28M images and the validation dataset contains approximately 50K images. For fairness, {the same training strategy as that employed by DeiT~\cite{touvron2021training} and PVT~\cite{wang2021pyramid} is followed to facilitate comparisons with other networks.} Specifically, {the} AdamW {is taken as the} parameter optimizer and the weight decay {rate} is set to 0.05. All {networks} are trained 300 epochs and the initial learning rate is set to 0.001 with {the} cosine decay {strategy}. We employ the same data augmentation techniques as {those used in} DeiT~\cite{touvron2021training}, including random flipping, random cropping, random erasing~\cite{zhong2020random}, CutMix~\cite{yun2019cutmix}, Mixup~\cite{zhang2017mixup} and label smoothing~\cite{szegedy2016rethinking}. Unless otherwise specified, the input image {resolutions} of {the} EViTs are all $224^2$ during the training process.

\begin{table*}[t]
	\caption{{Comparison among the} object detection (left group) and instance segmentation (right group) {performances achieved} on the COCO 2017 {validation} dataset. Each model is used as a visual backbone and then plugged into the RetinaNet~\cite{lin2017focal} and Mask R-CNN~\cite{he2017mask} frameworks.}
	\begin{center}
		\small\resizebox{1.0\linewidth}{!}{
			\begin{tabular}{l|ccccccc|ccccccc}
				\toprule[0.125em]
				\multirow{2}{*}{Backbone} & \multicolumn{7}{c|}{RetinaNet} & \multicolumn{7}{c}{Mask R-CNN}  \\
				& Params (M) &  $mAP$ & $AP_{50}$  & $AP_{75}$ & $AP_S$ &  $AP_M$ & $AP_L$ & Params (M) & $mAP^b$ & $AP^b_{50}$  & $AP^b_{75}$ & $mAP^m$ &  $AP^m_{50}$ & $AP^m_{75}$ \\
				\midrule
				ResNet-50~\cite{he2016deep} & 37.7 & 36.3 & 55.3 & 38.6 & 19.3 & 40.0 & 48.8 & 44.2 & 38.0 & 58.6 & 41.4 & 34.4 & 55.1 & 36.7 \\
				PVT-S~\cite{wang2021pyramid} & 34.2 & 40.4 & 61.3 & 43.0 & 25.0 & 42.9 & 55.7 & 44.1 & 40.4 & 62.9 & 43.8 & 37.8 & 60.1 & 40.3 \\
				FViT-Small~\cite{shi2024fvit} & 32.9 & 41.1 & 62.0 & 43.6 & 25.2 & 43.4 & 56.3 & 42.8 & 40.9 & 63.5 & 44.3 & 38.4 & 60.9 & 41.3 \\
				Swin-T~\cite{liu2021swin} & 38.5 & 42.0 & 63.0 & 44.7 & 26.6 & 45.8 & 55.7 & 47.8 & 42.2 & 64.6 & 46.2 & 39.1 & 61.6 & 42.0 \\
				ResT-Base~\cite{zhang2021rest} & 40.5 & 42.0 & 63.2 & 44.8 & 29.1 & 45.3 & 53.3 & 49.8 & 41.6 & 64.9 & 45.1 & 38.7 & 61.6 & 41.4 \\
				DAT-T~\cite{xia2022vision} & 38.0 & 42.8 & 64.4 & 45.2 & 28.0 & 45.8 & 57.8 & 48.0 & 44.4 & 67.6 & 48.5 & 40.4 & 64.2 & 43.1 \\
				CMT-S~\cite{guo2022cmt} & 44.0 & 44.3 & 65.5 & 47.5 & 27.1 & 48.3 & 59.1 & 45.0 & 44.6 & 66.8 & 48.9 & 40.7 & 63.9 & 43.4 \\
				PVTv2-B2~\cite{wang2022pvt} & 35.1 & 44.6 & 65.6 & 47.6 & 27.4 & 48.8 & 58.6 & 45.0 & 45.3 & 67.1 & 49.6 & 41.2 & 64.2 & 44.4 \\
				CrossFormer-S~\cite{wang2023crossformer++} & 40.8 & 44.4 & 65.8 & 47.4 & 28.2 & 48.4 & 59.4 & 50.2 & 45.4 & 68.0 & 49.7 & 41.4 & 64.8 & 44.6 \\
				\rowcolor{mygray}
				EViT-Small & {34.6} & {45.1} & {66.0} & {48.4} & {28.3} & {49.2} & {59.7} & {44.8} & {46.0} & {67.6} & {50.3} & {41.7} & {64.8} & {44.9} \\
				\midrule
				\midrule
				ResNet-101~\cite{he2016deep} & 56.7 & 38.5 & 57.8 & 41.2 & 21.4 & 42.6 & 51.1 & 63.2 & 40.4 & 61.1 & 44.2 & 36.4 & 57.7 & 38.8 \\
				PVT-M~\cite{wang2021pyramid} & 53.9 & 41.9 & 63.1 & 44.3 & 25.0 & 44.9 & 57.6 & 63.9 & 42.0 & 64.4 & 45.6 & 39.0 & 61.6 & 42.1 \\
				FViT-Base~\cite{shi2024fvit} & 50.3 & 42.2 & 63.9 & 44.8 & 25.8 & 45.3 & 58.5 & 60.3 & 42.4 & 64.9 & 46.4 & 39.6 & 62.2 & 42.8 \\
				Swin-S~\cite{liu2021swin} & 59.8 & 44.5 & 65.7 & 47.5 & 27.4 & 48.0 & 59.9 & 69.1 & 44.8 & 66.6 & 48.9 & 40.9 & 63.4 & 44.2 \\
				DAT-S~\cite{xia2022vision} & 60.0 & 45.7 & 67.7 & 48.5 & 30.5 & 49.3 & 61.3 & 69.0 & 47.1 & 69.9 & 51.5 & 42.5 & 66.7 & 45.4 \\
				ScalableViT-B~\cite{yang2022scalablevit} & 85.0 & 45.8 & 67.3 & 49.2 & 29.9 & 49.5 & 61.0 & 95.0 & 46.8 & 68.7 & 51.5 & 42.5 & 65.8 & 45.9 \\
				PVTv2-B3~\cite{wang2022pvt} & 55.0 & 45.9 & 66.8 & 49.3 & 28.6 & 49.8 & 61.4 & 64.9 & 47.0 & 68.1 & 51.7 & 42.5 & 65.7 & 45.7 \\
				CrossFormer-B~\cite{wang2023crossformer++} & 62.1 & 46.2 & 67.8 & 49.5 & 30.1 & 49.9 & 61.5 & 71.5 & 47.2 & 69.9 & 52.0 & 42.7 & 66.6 & 46.2 \\
				\rowcolor{mygray}
				EViT-Base & {52.5} & {46.5} & {67.5} & {49.8} & {29.4} & {51.3} & {62.1} & {63.4} & {47.5} & {68.8} & {52.3} & {43.1} & {66.3} & {46.3} \\    
				\bottomrule[0.125em]
		\end{tabular}}
	\end{center}
	\label{tab:coco2017}
\end{table*}

\noindent\textbf{Results.} Table~\ref{tab:class_results} shows the performance {achieved by the} EViTs {in} the ImageNet classification task. For ease of comparison, similar {networks are grouped} based on {their} model parameters and performance. {The experimental results show that the} EViTs achieve better accuracy and speed {trade-offs than the other models do} with similar model parameters. Specifically, EViT-Tiny and EViT-Small {achieve 79.9\% and 82.6\% classification {accuracies with} small model scales, respectively. Compared with the latest network, although EViT-Tiny is not as strong as RepViT-M1.5~\cite{wang2024repvit} {in terms of performance}, EViT-Small demonstrates competitive advantages at larger parameter scales.} {Compared with ConvNext-S~\cite{liu2022convnet}, Focal-S~\cite{yang2021focal} and ResT-Large~\cite{zhang2021rest},} EViT-Base {attains} impressive performance with the lowest computational cost. Specifically, EViT-Base achieves 83.9\% Top-1 accuracy with 7.2 GFLOPs, {representing a performance improvement of} 0.3\% over the three {above} mentioned methods while simultaneously reducing {the required} computational complexity by nearly 0.7 to 1.9 GFLOPs. At larger parameter scales, EViT-Large maintains significant competitive advantages over {the} other networks. In particular, for {a} fair comparison with {the} other {networks, two {image classification} experiments are conducted on EViT-Large {with} image sizes of $224^2$ and $256^2$.} {Under} the same settings, EViT-Large can {achieve 0.8\% and 0.6\% performance gains over PVTv2-B4~\cite{wang2022pvt} and ConvNext-B~\cite{liu2022convnet}, respectively.} When the input image size is set to $256^2$, the computational complexity and {number of} model parameters {required by} EViT-Large are only 13.7 GFLOPs and 61.9 M{,} respectively, {but it achieves} 84.9\% classification {accuracy}. {Additionally, {the} EViTs are compared with representative vision mamba models, including EfficientVMamba~\cite{pei2024efficientvmamba}, MSVMamba~\cite{shi2024multi}, and VMamba~\cite{liu2024vmamba}. {The numerical} results indicate that our method still demonstrates a competitive advantage over these linear attention models. It enhances the context-awareness of the model to better capture long-range dependencies.} As described, {the} EViTs exhibit significant competitive advantages, especially {in terms of their} low computational complexity and {high} scalability. {They} can be flexibly scaled to {form} smaller or larger models {according to} specific task requirements.

\subsection{Object Detection and Instance Segmentation}
\noindent\textbf{Settings.} In this section, object detection and instance segmentation experiments {are conducted} on the COCO 2017~\cite{lin2014microsoft} dataset {to evaluate the} EViTs. The COCO 2017 dataset contains 80 classes, 118k training images, 5k validation images and 20k test images. {Two} representative frameworks, RetinaNet~\cite{lin2017focal} and {the} Mask R-CNN~\cite{he2017mask} {are used to} evaluate the performance of {the} EViTs. Specifically, the EViTs are used as backbone {networks} and then plugged into the RetinaNet and {the} Mask R-CNN frameworks. Before {starting the} training {process}, {each backbone network is initialized by using the parameters determined during ImageNet-1k pretraining, whereas} the other layers are randomly initialized. For fairness, {the same settings as those of PVTv2~\cite{wang2022pvt} are followed.} AdamW is selected as {the} optimizer and the training schedule is set to 1 × 12 epochs. The weight decay {rate} and the initial learning rate are set to 0.05 and 0.0001, respectively.

\noindent\textbf{Results.} Table~\ref{tab:coco2017} shows the performance comparison {between the} EViTs {and} other backbone networks {in terms of} object detection and instance segmentation {tasks implemented} on {the} COCO 2017 dataset. In the RetinaNet framework, the mean Average Precision (mAP), Average Precision at 50\% and 75\% {Intersection over Union} (IoU) thresholds ($AP_{50}$, $AP_{75}$), and average precision across three object sizes (small, medium, and large, {denoted as} $AP_{S}$, $AP_{M}$, and $AP_{L}${, respectively}) are used as the evaluation metrics {for} model performance {assessments}. {The numerical results show} that {the} EViTs have significant competitive advantages {over the} other networks. Specifically, the average {accuracies} of EViT-Small and EViT-Base {are} at least 8\% higher than {those of} ResNet-50 and ResNet-101, and {they outperform} the advanced PVTv2-B2 and PVTv2-B3 by 0.5\% and 0.6\%, respectively. In the Mask R-CNN framework, the bounding box Average Precision ($AP_{b}$) and mask Average Precision ($AP_{m}$) at {the} mean and different IoU thresholds (50\%, 75\%) are used as the evaluation metrics. {These results also show that} EViT-Small and EViT-Base significantly outperform the other networks. Specifically, in {terms of the} $mAP^b$ and $mAP^m$ metrics, EViT-Small {exceeds} PVTv2-B2 {by} 0.7\% and 0.5\%, {while} EViT-Base {exceeds} PVTv2-B3 {by} 0.5\% and 0.6\% respectively.

\begin{table}[t]
	\caption{Comparison {among the} semantic segmentation {results obtained} with {a Semantic} FPN on ADE20K.}
	\begin{center}
		\resizebox{1.0\linewidth}{!}{
			\begin{tabular}{l|>{\centering\arraybackslash}m{1.5cm}|>{\centering\arraybackslash}m{1.5cm}|>{\centering\arraybackslash}m{1.5cm}}
				\toprule[0.125em]
				Backbone & Params (M) & FLOPs (G) & mIoU (\%) \\
				\midrule
				ResNet-50~\cite{he2016deep} & 28.5 & 45.6 & 36.7 \\
				PVT-S~\cite{wang2021pyramid} & 28.2 & {44.5} & 39.8 \\
				Swin-T~\cite{liu2021swin} & 32.0 & {46.0} & 41.5 \\
				LITv2-S~\cite{pan2022fast} & 31.0 & {41.0} & 44.3 \\
				ScalableViT-S~\cite{yang2022scalablevit} & 30.0 & {45.0} & 44.9 \\
				FaViT-B2~\cite{qin2023factorization} & 29.0 & {45.2} & 45.0 \\
				PVT v2-B2~\cite{wang2022pvt} & 29.1 & 45.8 & 45.2 \\
				CrossFormer-S~\cite{wang2023crossformer++} & 34.0 & 61.0 & 46.0 \\
				\rowcolor{mygray}
				EViT-Small & {28.6} & 44.8 & {46.1} \\
				\midrule
				\midrule
				ResNet-101~\cite{he2016deep} & 47.5 & 65.1 & 38.8 \\
				PVT-M~\cite{wang2021pyramid} & 48.0 & {61.0} & 41.6 \\
				Swin-S~\cite{liu2021swin} & 53.0 & {70.0} & 45.2 \\
				LITv2-M~\cite{pan2022fast} & 52.0 & {63.0} & 45.7 \\
				FaViT-B3~\cite{qin2023factorization} & 52.0 & {66.7} & 47.2 \\
				PVT v2-B3~\cite{wang2022pvt} & 49.0 & 62.4 & 47.3 \\
				CrossFormer-B~\cite{wang2023crossformer++} & 56.0 & 91.0 & 47.7 \\
				ScalableViT-S~\cite{yang2022scalablevit} & 79.0 & {70.0} & 48.4 \\
				\rowcolor{mygray}
				EViT-Base & {46.5} & 60.1 & {48.5}\\
				\bottomrule[0.125em]
		\end{tabular}}
	\end{center}
	\label{tab:ADE20K}
\end{table}

\subsection{Semantic Segmentation on ADE20K}

\noindent\textbf{Settings.} {Semantic segmentation experiments are conducted for the EViTs on the ADE20K~\cite{zhou2017scene} dataset. The} ADE20K dataset is widely used for semantic segmentation tasks and comprises 150 different semantic categories, with {approximately} 20K training images, 2K validation images, and 3K test images. To facilitate {comparisons} with other networks, {the EViTs are used as backbones and integrated into the Semantic FPN~\cite{kirillov2019panoptic}, enabling us to evaluate} the performance of {the} EViTs in semantic segmentation tasks. Specifically, {the same parameter settings as {those of} PVT~\cite{wang2021pyramid} are followed.} AdamW {is chosen as the} parameter optimizer, {and the learning rate is set to 0.0001.} The learning rate is decayed {by} following the polynomial decay schedule with {a} power {parameter of} 0.9, and the number of training iterations is 80k.

\noindent\textbf{Results.} 

Table~\ref{tab:ADE20K} shows the {results of a} performance comparison {between the} EViTs with {other backbone networks in terms of a} semantic segmentation {task implemented} on the ADE20K~\cite{zhou2017scene} dataset. In this experiment, the model parameters, computational complexity (FLOPs), and mean Intersection over Union (mIoU) are used as evaluation metrics. Specifically, EViT-Small and EViT-Base {achieve mean IoUs (mIoUs)} of 46.1\% and 48.5\%, respectively. Under similar parameter counts and FLOPs, the EViT-Small and EViT-Base outperform the LITv2~\cite{pan2022fast} and PVTv2~\cite{wang2022pvt} {approaches} by at least 1.8\% and 0.9\%, respectively. Additionally, other advanced networks, such as FaViT-B3~\cite{qin2023factorization}, CrossFormer-B~\cite{wang2023crossformer++}, and ScalableViT-S~\cite{yang2022scalablevit} are subjected to a more comprehensive comparison. Compared with these networks, EViT-Base leads in segmentation performance by 1.3\%, 0.8\%, and 0.1\%, respectively. The numerical results indicate that {the} EViTs demonstrate a significant competitive advantage over these networks in dense prediction tasks.

\subsection{Other Vision Transfer Learning Tasks}
\noindent\textbf{Settings.} In this section, other transfer learning experiments are conducted to evaluate the performance of {the} EViTs in different downstream vision tasks. These vision tasks consist of different application scenarios and datasets, including fine-grained visual classification (Standford Cars~\cite{krause20133d}, Oxford-102 Flowers~\cite{nilsback2008automated} and Oxford-IIIT-Pets~\cite{parkhi2012cats}), long-tailed classification (iNaturalist18~\cite{van2017inaturalist}, iNaturalist19~\cite{van2017inaturalist}) and superordinate classification tasks (CIFAR-10~\cite{krizhevsky2009learning}, CIFAR-100~\cite{krizhevsky2009learning}). The details of these datasets are listed in Table~\ref{tab:above_datasets}. 

\begin{table}[t]
	\caption{Details of {the employed} vision datasets. This table contains the number of classes, training images, and testing images {contained in} these datasets.}
	\begin{center}
		\resizebox{1.0\linewidth}{!}{
			\begin{tabular}{l|>{\centering\arraybackslash}m{1.0cm}|>{\centering\arraybackslash}m{1.25cm}|>{\centering\arraybackslash}m{1.25cm}}
				\toprule[0.125em]
				dataset & classes & train data & val data\\
				\midrule
				Standford Cars~\cite{krause20133d} & 196 & 8133 & 8041 \\
				Oxford-102 Flowers~\cite{nilsback2008automated} & 102 & 2040 & 6149 \\
				Oxford-IIIT-Pets~\cite{parkhi2012cats} & 37 & 3680 & 3669 \\
				\midrule
				iNaturalist18~\cite{van2017inaturalist} & 8142 & 437513 & 24426 \\
				iNaturalist19~\cite{van2017inaturalist} & 1010 & 265240 & 3003 \\
				\midrule
				CIFAR-10~\cite{krizhevsky2009learning} & 10 & 50000 & 10000 \\
				CIFAR-100~\cite{krizhevsky2009learning} & 100 & 50000 & 10000 \\
				\bottomrule[0.125em]
		\end{tabular}}
	\end{center}
	\label{tab:above_datasets}
\end{table}

\noindent\textbf{Results.} Before {beginning the} training {process}, we use the parameters {determined during} ImageNet-1K {pretraining} to initialize the EViT {backbones}, {while the} other layers are randomly initialized. Table~\ref{tab:transfer_learning} shows the {results of a} performance comparison between {the} EViTs and other backbone networks {in the aforementioned} vision tasks. The numerical results indicate that {the} EViTs demonstrate highly competitive performance. In {the} particular, EViTs {achieve} comparable or even superior performance {relative to that of} EfficientNet-B5 and EfficientNet-B7 at {a lower} computational cost. This demonstrates the superiority and generality of the EViTs based on {the bi-foveal eagle} vision designed in this paper.

\begin{table*}[t]
	\caption{{Results of a performance} comparison between {the} EViTs and other backbone networks {in a} fine-grained visual classification task, {a} long-tailed classification task and {a} superordinate classification task.}
	\centering
	\small\resizebox{1.0\linewidth}{!}{
		\begin{tabular}{l|c|c|c|c|c|c|c|c|c}
			\toprule [0.125em]
			Model & Params (M) & FLOPs (G) & Cars & Flowers & Pets & iNaturalist18 & iNaturalist19 & CIFAR-10 & CIFAR-100 \\
			\midrule
			Grafit ResNet-50~\cite{touvron2021grafit} & 25.6 & 4.1 & 92.5\% & 98.2\% & -\% & -\% & 75.9\% & -\% & -\% \\
			EfficientNet-B5$_{\uparrow456}$~\cite{tan2019efficientnet} & 28.0 & 10.3 & 93.9\% & 98.5\% & 94.9\% & -\% & -\% & 98.7\% & 91.1\% \\
			CeiT-S~\cite{yuan2021incorporating} & 24.2 & 4.8 & 94.1\% & 98.6\% & 94.9\% & 73.3\% & 78.9\% & 99.1\% & 90.8\% \\
			TNT-S$_{\uparrow384}$~\cite{han2021transformer} & 23.8 & 5.2 & -\% & 98.8\% & 94.7\% & -\% & -\% & 98.7\% & 90.1\% \\
			ViTAE-S~\cite{xu2021vitae} & {23.6} & 5.6 & 91.4\% & 97.8\% & 94.2\% & -\% & 76.0\% & 98.8\% & 90.8\% \\
			\rowcolor{mygray}
			{EViT-Small} & 24.9 & 4.1 &  93.6\% & 98.4\% &  95.0\% & 73.5\% &  79.1\% &  99.1\% &  91.2\% \\
			
			\midrule
			
			TNT-b$_{\uparrow384}$~\cite{han2021transformer} & 65.6 & 14.1 & -\% & 99.0\% & 95.0\% & -\% & -\% & 99.1\% & 91.1\% \\
			EfficientNet-B7$_{\uparrow600}$~\cite{tan2019efficientnet} & 64.0 & 37.2 &  94.7\% &  98.8\% &  95.4\% & -\% & -\% & 98.9\% &  91.7\% \\
			ViT-B/16$_{\uparrow384}$~\cite{dosovitskiy2020image} & 85.8 & 17.6 & -\% & 89.5\% & 93.8\% & -\% & -\% & 98.1\% & 87.1\% \\
			DeiT-B~\cite{touvron2021training} & 85.8 & 17.3 & 92.1\% & 98.4\% & -\% & 73.2\% & 77.7\% & 99.1\% & 90.8\% \\
			\rowcolor{mygray}
			{EViT-Base} & {42.8} & 7.2 &  94.7\% & 98.6\% & 95.2\% &  73.8\% &  79.5\% &  99.4\% &  91.7\% \\
			\bottomrule[0.125em]
	\end{tabular}}
	\label{tab:transfer_learning}
\end{table*}

\subsection{Robustness Evaluation}

\begin{table}[t]
	\caption{{The results of a performance comparison between the EViT-Small with other convolutional and vision transformer models on the ImageNet-C benchmark dataset. The mean corruption error (mCE) is used as metric. The smaller mCE value denotes higher model robustness under corruption.}}
	\begin{center}
		\small\resizebox{1.0\linewidth}{!}{
			\begin{tabular}{l|>{\centering\arraybackslash}m{1.5cm}|>{\centering\arraybackslash}m{1.5cm}|>{\centering\arraybackslash}m{1.35cm}}
				\toprule[0.125em]
				\multirow{1}{*}{Model} & \multirow{1}{*}{Params (M)} & \multirow{1}{*}{FLOPs (G)} & \multirow{1}{*}{ mCE } \\
				\midrule
				ResNet-50~\cite{he2016deep} & 25.6 & 4.1 & 76.7 \\
				EfficientNet-B4~\cite{tan2019efficientnet} & 19.3 & 4.4 & 71.1 \\
				RegNetY-4.0~\cite{radosavovic2020designing} & 20.6 & 4.0 & 68.7 \\
				Swin-T~\cite{touvron2021training} & 28.3 & 4.5 & 62.0 \\
				PVT-S~\cite{wang2021pyramid} & 24.5 & 3.8 & 66.9 \\
				DeiT-S~\cite{touvron2021training} & 22.1 & 4.6 & 54.6 \\
				RVT-S~\cite{mao2022towards} & 22.1 & 4.7 & {50.1} \\
				\rowcolor{mygray}
				{EViT-Small} & 24.9 & 4.1 & 52.4 \\
				\bottomrule[0.125em]
		\end{tabular}}
	\end{center}
	\label{tab:robustness}
\end{table}

\noindent\textbf{Settings.} {In this section, the robustness of the EViTs is evaluated on the ImageNet-C~\cite{hendrycks2019benchmarking} dataset, which consists of various types of image corruptions, categorized into noise, blur, weather, and digital. Each type of corruption has five severity levels, with varying intensities of degradation. The test results obtained on this benchmark are used to reflect the overall robustness of the developed networks. For performance evaluation purposes, the mean corruption error (mCE) is used as the metric. A smaller mCE indicates greater model robustness under corruptions.}

\noindent\textbf{Results.} {Table~\ref{tab:robustness} shows the experimental results produced by EViT-Small on the ImageNet-C benchmark dataset. Several convolutional neural networks and vision transformers used for comparison, where RVT is a specially designed vision transformer network that achieves high robustness. The numerical results show that the EViT-Small achieves similar performance to that of RVT-S in the robustness tests. It demonstrates good robustness under various types of corruptions and further validating the practicality of the proposed method. Although our focus is not on designing a robust visual transformer akin to RVT, several design principles derived from RVT are naturally incorporated into the EViTs, such as the convolutional stem, position embedding, pyramidal network architecture, and local information exchange in the feedforward network. These key designs ensure that the EViTs exhibit robust performance under various types of image corruption.}

\begin{figure}[t]
	\centering
	\small\resizebox{1.0\linewidth}{!}{
		\begin{tabular}{c}
			\hspace{-3mm}\includegraphics[width=0.515\textwidth]{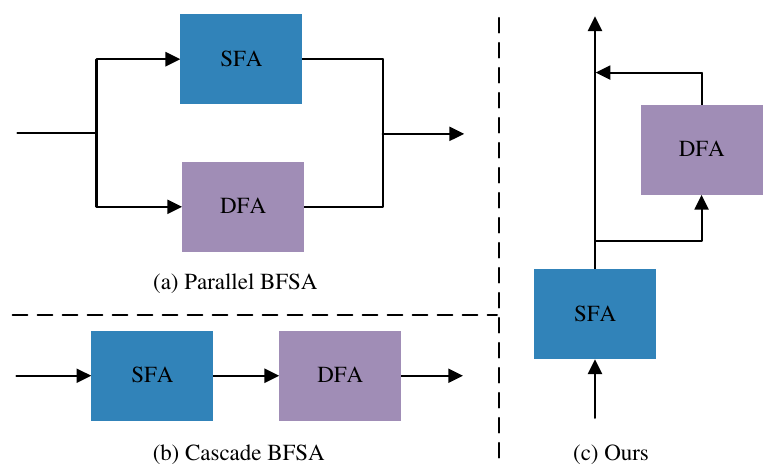}
	\end{tabular}}
	\caption{The three connection methods {used by the} BFSA.}
	\label{fig:BFSA_Archs}
\end{figure}

\subsection{Ablation Study}

\noindent\textbf{Settings.} In this section, {the ablation experiments are conducted on the} ImageNet-1K~\cite{russakovsky2015imagenet} dataset to demonstrate {the effectiveness of the BFVI connection structure and the BFFN. Specifically, the ablation study uses EViT-Base as the baseline. The training strategy outlined} in Section 4.1 {is followed}.

\noindent\textbf{Effectiveness analysis of the BFVI.} 

The Bi-Fovea Self-Attention (BFSA) {and Bi-Fovea Feedforward Network (BFFN) are the main contributions} of our work. {They are the basic components used to building the EViTs, ensuring that these EViTs can achieve competitive performance in various vision tasks. We attribute the advantages of the EViTs to the unique visual structure derived from eagle vision, which is referred to as Bi-Foveal Visual Interaction (BFVI).} {Taking BFSA as an example,} Figure~\ref{fig:BFSA_Archs} (c) {illustrates} the unique connection approach {employed by} the Shallow Fovea Attention (SFA) and Deep Fovea Attention (DFA) in {the BFSA via} simplification. The BFSA {does} not simply {connect} the SFA and DFA in parallel or cascade, it is more {similar to} the combination of them. {Therefore, the ablation studies are first conducted on BFVI to demonstrate the effectiveness of this connection approach.} 

In the implementation details, {the Parallel BFSA and Cascade BFSA are implemented by connecting the SFA and DFA in parallel and in a cascaded manner, respectively. Similarly, a Parallel BFFN and a Cascaded BFFN are also constructed, and they} are used {to conduct a comparison with the proposed BFSA and BFFN.} Table~\ref{tab:ablation_study} shows the performance comparison {conducted among} three connection {approaches} for {the} BFSA and BFFN. {Note that although these three connection modes have the same number of parameters and computational complexity, our BFVI connection approach achieves superior performance.} Specifically, {the} BFSA outperforms {the} Parallel BFSA and Cascade BFSA by 1.4\% and 0.5\%{, respectively, in terms of the Top-1 classification accuracy metric}. {The performance of {the} BFFN exceeds that of {the} Parallel BFFN and Cascade BFFN by 0.7\% and 0.2\%, respectively. This demonstrates that the BFVI combines the advantages of parallel and cascaded patterns, and can attain more} competitive performance in vision tasks.

\noindent\textbf{Effectiveness analysis of BFFN.} 

\begin{table}[t]
	\caption{Results of the ablation experiments {conducted on the} BFSA and BFFN.}
	\begin{center}
		\resizebox{1.0\linewidth}{!}{
			\begin{tabular}{l|c|c|c}
				\toprule[0.125em]
				\multirow{1}{*}{Method} & FLOPs (G)& Params (M)& Top-1 Acc (\%).\\
				\midrule
				+ Parallel BFSA & 7.16 & 42.8 & 82.5 \\
				+ Cascade BFSA & 7.16 & 42.8 & 83.4 \\
				+ BFSA (Ours) & 7.16 & 42.8 & 83.9 \\
				\midrule
				+ Parallel BFFN & 7.16 & 42.8 & 83.2 \\
				+ Cascade BFFN & 7.16 & 42.8 & 83.7 \\
				+ BFFN (Ours) & 7.16 & 42.8 & 83.9 \\
				\midrule
				+ FFN & 6.95 & 42.0 & 82.8 \\
				+ CFFN & 7.03 & 42.4 & 83.5 \\
				+ BFFN & 7.16 & 42.8 & 83.9 \\
				\bottomrule[0.125em]
		\end{tabular}}
	\end{center}
	\label{tab:ablation_study}
\end{table}

To demonstrate the effectiveness of {the} BFFN, {an ablation experiment is conducted in this section.} Specifically, the Feed Forward Network (FFN) {obtained} from {the} ViT~\cite{dosovitskiy2020image} and the Convolutional Feed Forward Network (CFFN) from PVT~\cite{wang2022pvt} are selected as the {controls}. The BFFN is {compared} with the FFN and the CFFN, respectively. Table~\ref{tab:ablation_study} presents the performance comparison {conducted between the} BFFN, FFN and CFFN. {The numerical results show that {the} BFFN outperforms {the} FFN and CFFN by 0.9\% and 0.4\%, respectively, while {its} computational cost remains negligible.} This {finding} demonstrates that {the} BFFN more efficiently complements the local detail information {contained} in the feed forward network, which is critical for computer vision tasks.

\section{Conclusion}
 
{In this paper, a} novel Bi-Fovea Self-Attention (BFSA) and Bi-Fovea Feedforward Network (BFFN) {are proposed.} Their core {ideas are} derived from the unique bi-fovea structure of eagle {eyes}, {which is referred to as Bi-Foveal Visual Interaction (BFVI).} {The} BFSA and BFFN can facilitate networks to model the global feature dependencies of images while extracting fine-grained feature representations of {the} targets. Additionally, a Bionic Eagle Vision (BEV) block {is designed {on the basis of} the BFSA and BFFN.} This BEV block combines the advantages of convolutions and transformers. Furthermore, a general {family of pyramidal} vision backbone {networks} called Eagle Vision Transformers (EViTs) is constructed by stacking BEV blocks. Experimental results show that {the} EViTs demonstrate excellent performance in image classification, object detection and semantic segmentation tasks. {In the future, we will further explore the advantages of combining eagle vision with vision transformers and conduct studies on linear attention models to {increase} the applicability and interpretability of {the} EViTs.}
 
{\small
	\bibliographystyle{IEEEtran}
	\bibliography{egbib}
}

\end{document}